\documentclass[10pt,twocolumn,letterpaper]{article}

\usepackage[pagenumbers]{cvpr} 

\usepackage{bbm}
\usepackage{soul}
\usepackage{pifont}
\usepackage{dsfont}
\usepackage{xcolor}
\usepackage{comment}
\usepackage{amsmath}
\usepackage{amssymb}
\usepackage{booktabs}
\usepackage{enumitem}
\usepackage{multirow}
\usepackage{graphicx}
\usepackage{colortbl}
\usepackage{adjustbox}
\usepackage{threeparttable}

\definecolor{mygray}{gray}{.85}
\newcommand{\xmark}{\ding{55}}%
\usepackage[pagebackref,breaklinks,colorlinks]{hyperref}

\usepackage[capitalize]{cleveref}
\crefname{section}{Sec.}{Secs.}
\Crefname{section}{Section}{Sections}
\Crefname{table}{Table}{Tables}
\crefname{table}{Tab.}{Tabs.}

\def\cvprPaperID{ID} 
\def\confName{CVPR}
\def\confYear{2022}

\begin{document}

\title{Self-supervised Tumor Segmentation through Layer Decomposition}

\author{Xiaoman Zhang\textsuperscript{1,2} \quad Weidi Xie\textsuperscript{3} \quad Chaoqin Huang\textsuperscript{1,2} \quad Yanfeng Wang\textsuperscript{1,2}\\
\quad Ya Zhang\textsuperscript{1,2} \quad Xin Chen\textsuperscript{4} \quad Qi Tian\textsuperscript{4}\\[3pt]
\textsuperscript{1}Cooperative Medianet Innovation Center, Shanghai Jiao Tong University \\
\textsuperscript{2}Shanghai AI Laboratory \quad
\textsuperscript{3}VGG, University of Oxford\\
\textsuperscript{4}Huawei Cloud \& AI\\[3pt]
\tt\small{\{xm99sjtu, huangchaoqin, wangyanfeng, ya\_zhang\}@sjtu.edu.cn, weidi@robots.ox.ac.uk}\\
\tt\small{chenxin061@gmail.com, tian.qi1@huawei.com }
}

\maketitle

\begin{abstract}
In this paper, 
we target self-supervised representation learning for zero-shot tumor segmentation.
We make the following contributions:
First, we advocate a zero-shot setting, 
where models from pre-training should be directly applicable for the downstream task, 
without using any manual annotations. 
Second, we take inspiration from ``layer-decomposition",
and innovate on the training regime with simulated tumor data.
Third, we conduct extensive ablation studies to analyse the critical components in data simulation,
and validate the necessity of different proxy tasks.
We demonstrate that, with sufficient texture randomization in simulation,
model trained on synthetic data can effortlessly generalise to segment real tumor data.
Forth, our approach achieves superior results for zero-shot tumor segmentation on different downstream datasets, BraTS2018 for brain tumor segmentation and LiTS2017 for liver tumor segmentation. While evaluating the model transferability for tumor segmentation under a low-annotation regime, 
the proposed approach also outperforms all existing self-supervised approaches, 
opening up the usage of self-supervised learning in practical scenarios.
\end{abstract}

\section{Introduction}
\label{sec:introduction}

\begin{figure}[t]
\centering{\includegraphics[width=.95\linewidth]{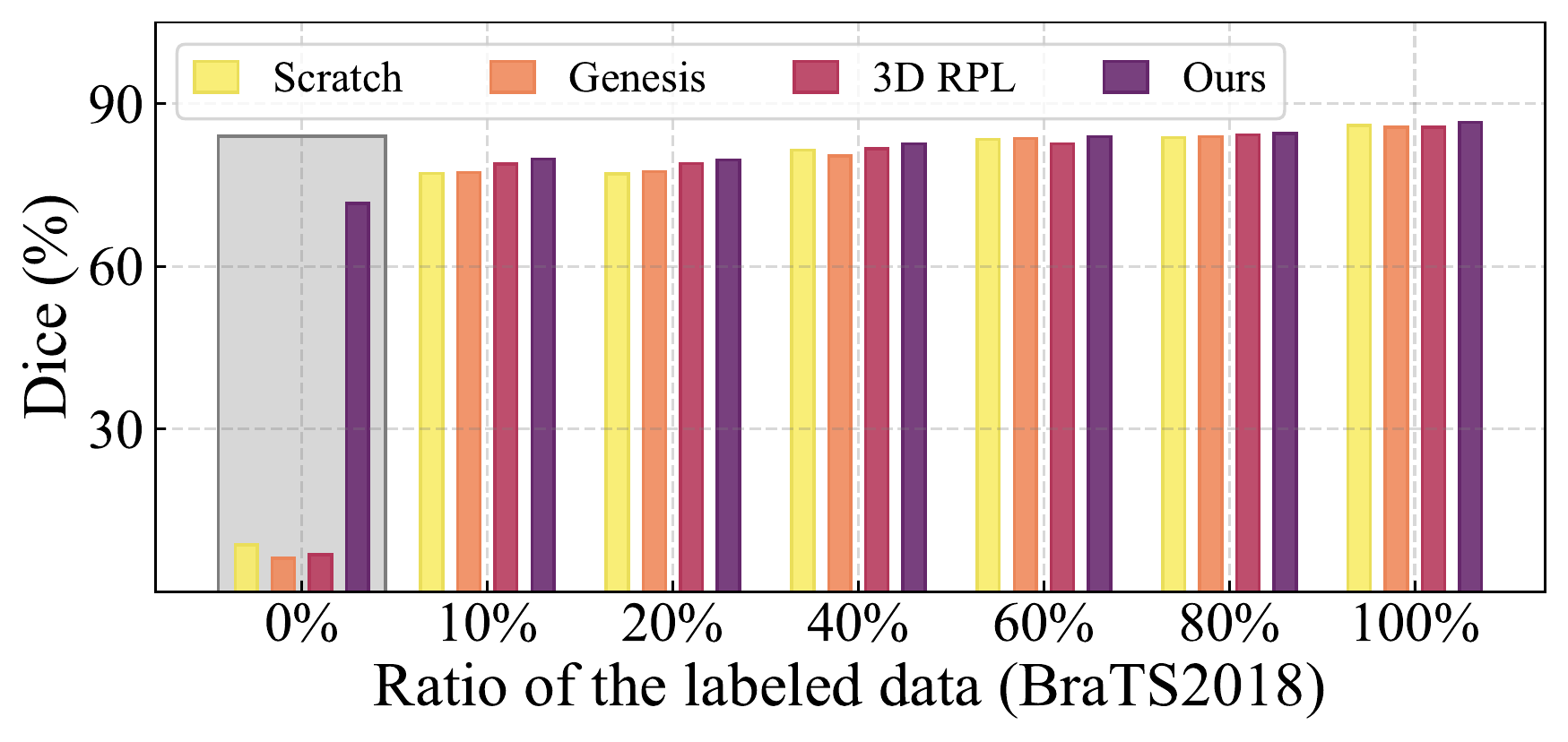}} \\
\centering{\includegraphics[width=.95\linewidth]{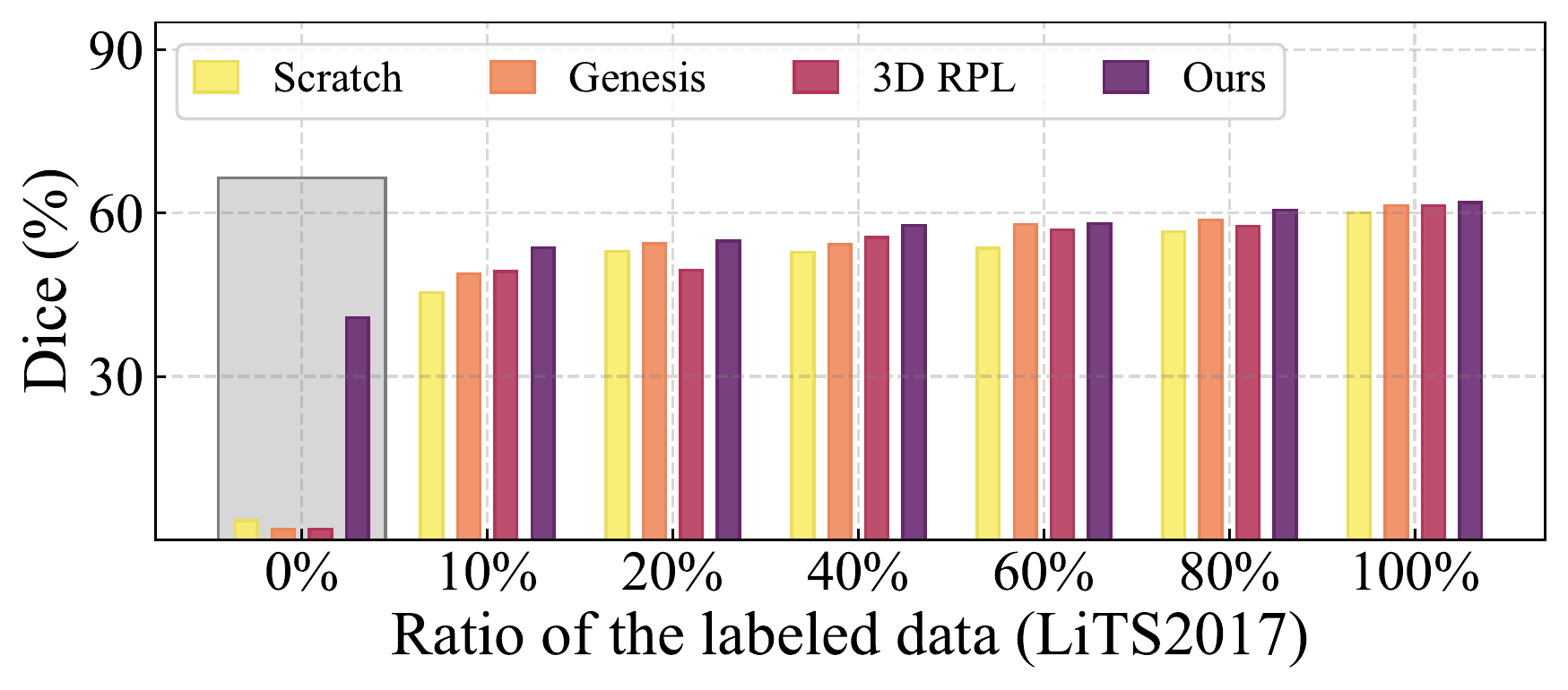}} 
\vspace{-8pt}
\caption{Comparison with SOTA self-supervised tumor segmentation approaches.
In contrast to the pretrain-finetune paradigm, 
our proposed model supports zero-shot segmentation, 
{\em i.e.}~models trained with self-supervised learning can directly generalise to the downstream task. When a few annotations are presented, 
the model can be finetuned to support efficient transfer.\vspace{-.4cm}}
\label{fig1}
\end{figure}

In medical image analysis, 
segmentation has been treated as one of the most important intermediate representations for obtaining reliable clinical measurements,
for example, size, shape, location, 
texture characteristics, or even growth rate of certain structures or anatomy, {\em etc.}
Such measurements can assist clinicians to make assessments of corresponding diseases
and better treatment plans.

In this paper, we consider the problem of {\em segmenting tumors} in 3D volumes.
In the recent literature, 
remarkable progress has been made with the use of deep neural networks,
however, 
these successful approaches often fall into the paradigm of supervised learning, 
requiring a large number of manually annotated volumes for training.
On tumor segmentation, 
acquiring the pixelwise annotations can often be a prohibitively expensive process,
due to the large variability in tumors' shape, size and location~\cite{ISIN2016317},
posing limitations on the scalability of the supervised training paradigm.

To alleviate the dependency on manual annotations,
the research community has recently shifted to a new learning paradigm,
namely, self-supervised learning~(SSL), 
with the goal of learning effective visual representation from raw data.
Generally speaking, 
the existing self-supervised approaches~\cite{Zhuang2019SelfSupervisedFL,Chen2019SelfSupervisedLF,Bai2019SelfSupervisedLF} 
in medical image analysis have predominantly followed a two-stage training paradigm, 
where the model is firstly pre-trained on unlabelled data by solving certain proxy tasks, and further adapted to the downstream segmentation task through {\em
fully supervised fine-tuning}, {\em i.e.}~the manual annotations remain prerequisite.

As an alternative, 
we consider a self-supervised learning approach for {\em zero-shot} tumor segmentation, 
as well as efficient transfer when annotations are provided.
To achieve such goal, 
the proxy task should ideally be defined in a way that minimises the generalisation gap with the considered downstream task, that is to say,
we would like the tumor segmentation mask to be somehow incorporated into the procedure for solving the proxy task.
Taking the observation that tumors are often characterised independently to their contexts, 
we propose a generative model that treats each sample as a linear composition between tumor and normal organ, 
with the segmentation mask naturally emerging from such decomposition.
In fact, such idea can be dated back to the classical layered representation in computer vision~\cite{Adelson341105},
though one crucial difference exists, that is, we do not favor high-quality reconstructions,
only the weights for alpha blending matters, 
which determine each pixel's assignment to tumor or organ.

In this paper, 
we innovate on the self-supervised training mechanism for layer decomposition, 
specifically, we propose a scalable pipeline for generating synthetic training samples by blending artificial ``tumors'' into normal organ,
and task a network to invert such generation process.
Once trained on synthetic data, the model can directly infer tumor masks on real samples,
{\em without} any fine-tuning whatsoever, advocating a Sim2Real learning regime.
In addition, whenever manual annotations are available,
performance can further be boosted, demonstrating superior {\em transferability},
as shown in Figure~\ref{fig1}.

To summarise, we make the following contributions:
{\em First}, 
we consider the problem of tumor segmentation under a zero-shot setting,
where models trained with self-supervised learning can be directly applicable for downstream task.
{\em Second}, 
we take inspiration from ``layer-decomposition", 
and innovate on the training regime with simulated tumor data.
{\em Third},
we conduct extensive ablation studies to analyse the critical components in data simulation, and validate the necessity of proposed proxy tasks,
demonstrating that, with sufficient texture randomization in simulation, 
models trained on synthetic data can effortlessly generalise to segment real tumor data.
Overall, 
our approach achieves superior results for zero-shot tumor segmentation on different downstream datasets, BraTS2018 for brain tumor segmentation and LiTS2017 for liver tumor segmentation. While evaluating the model transferability for tumor segmentation under a low-annotation regime, the proposed approach also outperforms all existing self-supervised approaches, opening up the usage of self-supervised learning in practical scenarios.

\begin{figure*}[t]
\centering{\includegraphics[width=.9\textwidth]{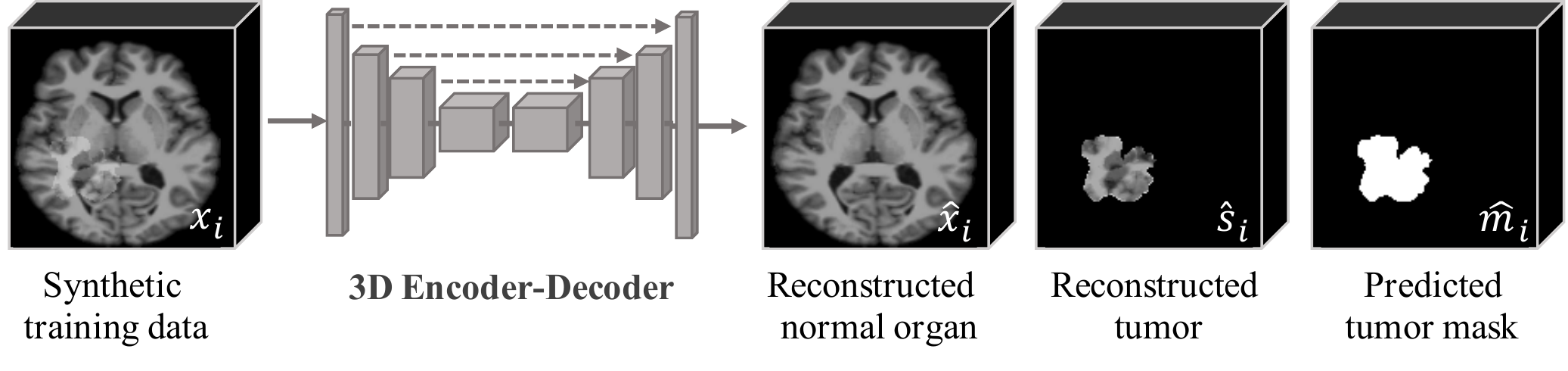}}
\vspace{-8pt}
\caption{\textbf{Illustration of the general idea.} 
Our model takes the synthetic training data $x_i$ as input, 
and outputs the reconstruction of normal organ $\hat{x}_{i}$, 
tumor $\hat{s}_{i}$ and the prediction of tumor mask $\hat{m}_{i}$. 
During inference, our pre-trained model can directly generalise to real tumor data,
{\em i.e.}~the prediction of tumor mask is treated as the segmentation result.
\vspace{-.2cm}}
\label{fig2}
\end{figure*}

\section{Related Work}
\label{sec2}
\par{\noindent \bf Tumor Segmentation.}
Automatic tumor segmentation in medical imaging is a very challenging task due to the fuzzy boundary and high variation in size, shape and location. 
In the recent literature, 
the progress has been mainly driven by the success of deep neural networks~\cite{Xiaomeng2018,Parvez2021,Isensee2021,Yeung2021}.
Two protocols have attracted increasing attention from the community,
namely, 
{\bf semi-supervised} tumor segmentation~\cite{CHEPLYGINA2019280,Shaobo2019,Shuai2019} 
and {\bf unsupervised} tumor segmentation~\cite{BAUR2021101952, marimont2021implicit, Dey2021ASCNetA}.
Specifically, in semi-supervised tumor segmentation, 
only a subset of training samples are manually annotated,
and the goal is to improve the model's performance with large amounts of unlabeled data.
In the unsupervised tumor segmentation setting, 
previous methods treat this problem as anomaly detection,
with the underlying assumption being that, 
images from healthy patients tend to have similar patterns, 
and tumors are therefore outliers.
In general, 
there are two lines of research, 
{\em e.g.}~generative or discriminative training.
In the former scenario, 
autoencoders~\cite{Huang9311201, Zong2018DeepAG} are trained to reconstruct the normal samples, 
with anomalies being the regions with high reconstruction error.
As an alternative, discriminative models directly train models with synthetic data, 
and output the tumor segmentation mask~\cite{to2021selfsupervised,Tan2021DetectingOW,Perez2021self}. 
In contrast, our proposed approach takes the benefits of both worlds, not only learning the representation of the normal image through reconstruction, 
but also predicting the segmentation mask of the detected anomalies. \\[-8pt]

\par{\noindent \bf Self-supervised Visual Representation Learning. }
Self-supervised learning~\cite{doersch2015unsupervised,noroozi2017unsupervised,Caron_2018_ECCV,Gidaris2018iclr,pathak2016context,hjelm2019learning,Feng8953870, Aaron2019cpc,Kaiming9157636,Chen2020ASF}
aims to learn effective visual representation from raw data {\em without} using manual annotations.
In general, 
most of existing frameworks in medical imaging have been migrated from computer vision \cite{NEURIPS2020_d2dc6368,Zhuang2019SelfSupervisedFL,Chen2019SelfSupervisedLF,Zhang2021sar}, 
and follow a two-stage training procedure, 
first, 
a model is trained by solving the predefined proxy task on a large-scale dataset, 
the learned representation is further fine-tuned for the downstream tasks with annotated data.
Note that, in these aforementioned approaches, 
proxy tasks are defined with limited prior knowledge from downstream task, 
thus, manual annotations are still required for fine-tuning.
In contrast to these two-stage training, we target direct generalisation, 
by that we mean, once trained to solve the proxy tasks, 
the self-supervised model should be able to segment tumors {\em without} fine-tuning on any annotated data, {\em i.e.}~zero-shot tumor segmentation.
In addition, when a small number of annotated data is available, 
our proposed model can be fine-tuned to further boost the segmentation performance. \\[-8pt]

\noindent \textbf{Layered Representations.}
Layered representations were first proposed to decompose videos into layers with simpler motions in~\cite{Wang334981}.
Since then, 
layer representations of images and videos have been widely adopted in computer vision, 
for inferring synthesis depth~\cite{Adelson341105,Brostow791190,Srinivasan2019,lsiTulsiani18}, 
separating reflections~\cite{alayrac2019visual,lu2020,Lu_2021_CVPR}, 
and performing foreground/background estimation~\cite{DoubleDIP,ke2021bcnet,Kumar1541236,Yang_2021_ICCV}.
All previous approaches operate on natural images or videos, 
in contrast, we focus on the scenario of medical imaging, 
treating the tumor image as a composition of normal organ and tumor, 
and tackle the tumor segmentation problem through layer decomposition.\\[-8pt]

\noindent \textbf{Sim2Real Transfer.} 
In many computer vision and robotic tasks, 
synthetic datasets prove to be a useful source for training,
with an infinite amount of training data.
This is especially true when the ground-truth is difficult or impossible to acquire at scale, for example, optical flow~\cite{Sintel2012, Mayer2016}, 
and text detection and recognition~\cite{Jaderberg2014SyntheticDA, Gupta2016synthetic}, 
pose estimation~\cite{Doersch2019Sim2realTL}. 
To improve the generalisation ability in Sim2Real transfer,
Domain Randomization~\cite{tobin2017domain, random2018Peng} is often adopted.
In this work, 
we adopt similar high-level idea,
{\em i.e.}~to train zero-shot tumor segmentation model on artificially simulated tumor data.
We demonstrate that,
with sufficient texture randomization in the generation process, 
the model demonstrates direct generalisation to segmenting real tumor data.

\section{Methods}
\label{sec3}
In this paper, 
we propose a self-supervised approach to learn effective representation for tumor segmentation.
Our key observation is that tumors are usually characterised by a region with different intensities or textures to the nearby tissue, {\em i.e.}~independent to the surrounding contexts.
We therefore treat each tumor sample as a composition of tumor and normal organ,
and train a model to separate the sample into two ``layers'', 
namely, tumor and health organ, along with a mask for linear blending.

Formally, given a set of 3D volume samples,  
$\mathcal{D} =\{ x_1, x_2, \dots, x_N\}$,
where $ x_i \in \mathbb{R}^{H \times W \times D \times 1}$, 
the goal is to train a computational model that outputs three elements, 
such that, their linear combinations can cover the input sample~(Eq.~\eqref{eq2}):
\begin{align}
\label{eq1}
  \hat{x}_i \text{,\hspace{5pt}} \hat{s}_i \text{,\hspace{5pt}} \hat{m}_i =  \Phi(x_i) 
\end{align}
\\[-1.2cm]
\begin{align}
\label{eq2}
      (\mathbf{1}- \hat{m}_i) \cdot \hat{x}_i + \hat{m}_i \cdot \hat{s}_i= x_i
\end{align}

\noindent 
Here, $\Phi(\cdot)$ denotes the parameterised model, 
$x_i$ refers to the input tumor sample, 
$\hat{x}_i$ is the reconstruction of normal organ, 
$\hat{s}_i$ is the reconstructed tumor, 
and as a side product, $\hat{m}_i$ ends up being the mask for tumor segmentation.
In practise, 
given the strong representation capability of deep neural networks, 
na\"ively training $\Phi(\cdot)$ with real tumor data may lead to trivial solutions, 
for example, $\hat{x}_i = x_i$, $\hat{m}_i = \mathbf{0}$,
as we have neither the corresponding image of organ, nor the tumor mask. 
We thus innovate on the training regime by using synthetic data,
as detailed in the following sections. 

\begin{figure*}[!htb]
\centering{\includegraphics[width=.9\textwidth]{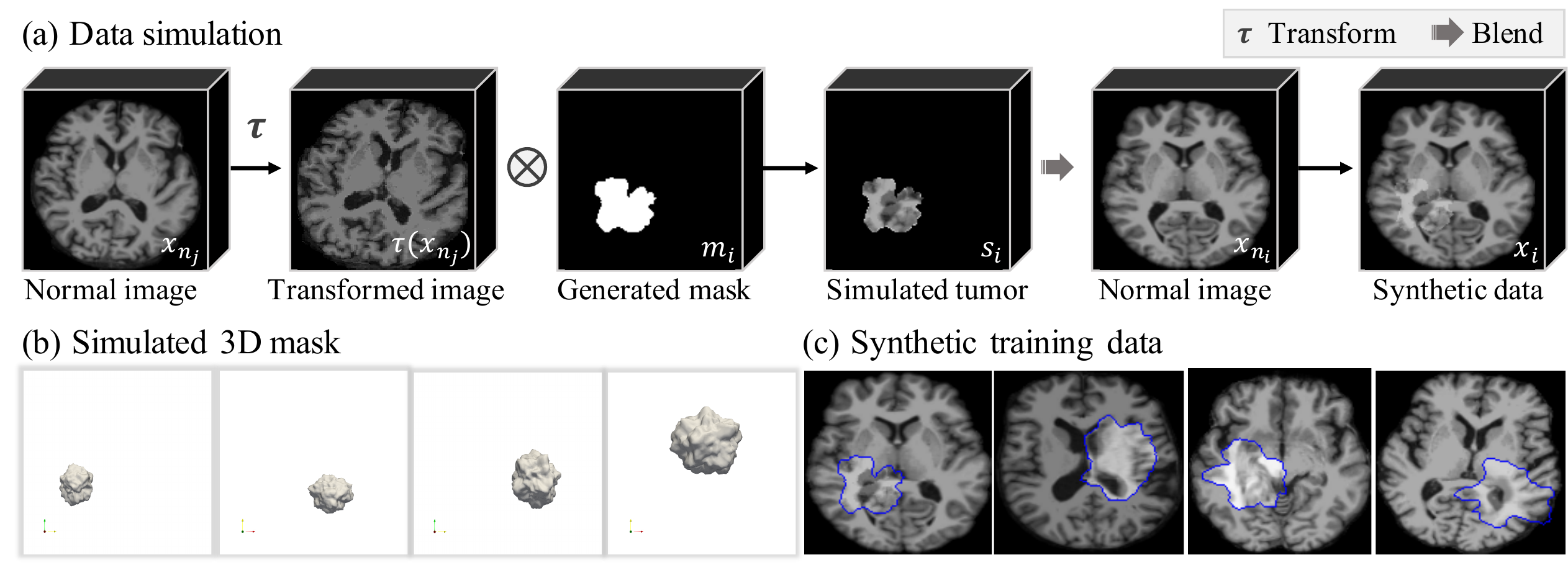}}
\vspace{-0.2cm}
\caption{(a) Data simulation pipeline. We first simulate the tumor with a transformed image and a generated 3D mask, which provide texture and shape respectively. Then the synthetic data is composed by blending the simulated tumor into the normal image.
(b) Visualisation of the simulated 3D masks with various shapes, sizes and locations. 
(c) Visualisation of synthetic images. Blue contour denotes the boundary of the simulated tumor.  \vspace{-0.3cm}} 
\label{fig3}
\end{figure*}

\subsection{Data Simulation}
\label{sec3.1}
In medical imaging, 
tumors can vary in size from a tiny nodule to a large mass with different shapes, 
and often appear distinct from the surrounding normal tissues. 
This motivates the idea of simulating tumor images by blending artificial anomalies 
into the normal organ, as illustrated in Figure~\ref{fig3} (a). 
In order to imitate a tumor sample, 
we randomly crop a region from one image and blend it into another one.
In the following paragraphs, 
we will elaborate the two essential elements for such data simulation,
namely, shape generation, texture randomization. \\[-4pt]

\par{\noindent \bf Shape Simulation. \hspace{5pt}}
To simulate the shape of tumors, 
we construct 3D masks of random size and position with non-smooth boundaries, 
as shown in Figure.~\ref{fig3} (b). 
We first pick a random point in the region-of-interest, {\em i.e.}~on the target organ, 
as the center coordinate of the mask.
Then, we generate a geodesic polyhedron by subdividing the edges of an icosahedron 
and projecting each vertex onto a parametric sphere of unit radius centered on the selected point.
We refer the readers to~\cite{Perez2021self} for more details.
The polyhedron is further perturbed with simplex noise to create a non-smooth surface that more closely resembles the appearance of the real tumor.
In addition, we apply transformations like random rotation and scaling to modify its orientation and volume.
Finally, the generated polyhedron is converted to a 3D binary mask~(${m}_i$) that indicates the tumor's shape and position.
Next, we will describe the detail to texture it.\\[-4pt]

\par{\noindent \bf Texture Simulation. \hspace{5pt}}
To simulate the texture of tumors, 
we only assume the existence of a set of images of normal organ,
{\em e.g.}~$\mathcal{D}_{\text{normal}} = \{x_{n_1}, x_{n_2}, \dots\}$.
In the following, 
we describe the tumor texture simulation process for a sampled normal image $x_{n_i}$,
with the simulated tumor mask ${m}_i$ from previous shape simulation.
Specifically, we construct the texture by borrowing from another randomly selected normal image $x_{n_j}$:
\begin{align}
    s_i = m_i \cdot \tau(x_{n_j}) \label{eq3}
\end{align}
where $\tau(\cdot)$ denotes a transformation function, 
and $s_i$ ends up being the synthetic tumor with texture. 

We adopt three commonly-used transformations as $\tau(\cdot)$, 
namely, linear transformation, 
elastic deformation and Gaussian blur, 
together controlling the difficulty level for the self-supervised proxy task.
Linear transformation maps the intensity of the selected normal image $x_{n_j}$ to the desired range,
which can be formulated as: 
\begin{align}
    \tau_{LT}(x_{n_j}) = r \cdot \frac{\mathrm{Mean}(x_{n_i}) }{\mathrm{Mean}(x_{n_j})} \cdot x_{n_j} \label{eq4}
\end{align}
where $r \in (a,b)$ is the normalisation ratio, 
$\mathrm{Mean}(\cdot)$ denotes the average intensity. 
This simple transformation is the key to simulating anomalies,
for example, if the tumor intensity is very different from the normal organ, 
the task can be trivially solved, while if the intensity is very close to the organ, 
the model would struggle to separate them, 
as it violates the assumption that tumor and organ tend to be visually {\em independent}.
The Gaussian blur aims to remove the high-frequency signals, 
such as sharp boundaries, strong textures,
and elastic deformation further augments the shape of tumors.
We refer the reader to supplementary material for more ﬁgures and details on the generation procedure. \\[-8pt]

\par{\noindent \bf Composition. \hspace{5pt}}
With the generated tumor ${s}_{i}$,
we can thus generate the training sample $x_i$ 
by blending the tumor into normal image $x_{n_i}$ with an alpha mask,
{\em i.e.}~processing the generated binary tumor mask with a variable mixing parameter $\alpha \in [0,1]$:
\begin{align}
 x_{i} = (\mathbf{1} - \alpha \cdot {m}_{i}) \cdot x_{n_i} + \alpha \cdot {m}_{i}\cdot {s}_{i} \label{eq5}
\end{align}

Note that, to avoid abusing the notations, 
we here use the same ones as in Eq.~\eqref{eq1} and Eq.~\eqref{eq2}, 
however, here $x_i, s_i$ and $m_i$ come from simulation with ground truth being available.
Till here,
it is trivial to train the decomposition model~{$\Phi(\cdot)$} with supervision on all the output elements.
Amazingly, as in Section~\ref{sec5.1}, 
we show that, 
simply training on these synthetic data already enables the model to directly generalise towards real data, with the inferred map as tumor segmentation mask.
In the next section, we will detail the training procedure. 

\subsection{Learning to Decompose}
\label{sec3.2}
Given the synthetic dataset described in the previous section, 
{\em e.g.}~$\mathcal{D_{\text{synthetic}}} = \{d_1, d_2, \dots, d_n\}$,
where $d_i = \{x_i, x_{n_i}, s_i, m_i\}$, 
refers to the set consisting synthetic tumor sample, 
normal organ image, simulated tumor and mask respectively.
We propose to train a network $\Phi(\cdot)$ that 
separately reconstructs the normal organ, tumor, and the segmentation mask,
{\em i.e.}~to invert the data simulation process~(Figure~\ref{fig3} (a)), 
recalling Eq.~\eqref{eq1}:
\begin{align*}
  \hat{x}_i \text{,\hspace{5pt}} \hat{s}_i \text{,\hspace{5pt}} \hat{m}_i =  \Phi(x_i) 
\end{align*}

In detail, we adopt standard L1 loss for the reconstruction of organ and tumor,
and Dice loss is used for optimising the mask prediction. 
The loss functions can be expressed as follows:
\begin{align}
    &\mathcal{L}_0 =  |\hat{x}_{i} - x_{n_i}| \label{eq6}\\ 
    &\mathcal{L}_1 = |\hat{s}_i - {s}_{i}| \label{eq7}\\
    &\mathcal{L}_2 = 1- \frac{2 \cdot |\hat{m}_i \cap {m}_i|}{|\hat{m}_i| + | {m}_i|} \label{eq8}\\
    &\mathcal{L}_3 =  |\left ((\mathbf{1} - \alpha \cdot \hat{m}_{i}) \cdot \hat{x}_i + \alpha \cdot \hat{m}_i \cdot \hat{s}_{i} \right) - {x}_{i}| \label{eq9}\\
    & \mathcal{L}_{\text{total}} = \lambda_0 \cdot \mathcal{L}_0 + \lambda_1 \cdot \mathcal{L}_1  + \lambda_2 \cdot \mathcal{L}_2 +  \lambda_3 \cdot \mathcal{L}_3 \label{eq10}
\end{align}
where $\lambda_0$, $\lambda_1$, $\lambda_2$ and $\lambda_3$ are parameters weighting the importance of each optimisation function. 
Note that, we directly use the original mixing parameter $\alpha$ to generate the reconstruction of blended input,
as shown in Eq.~\eqref{eq9},
such operation effectively forces the predicted segmentation mask~($m_i$) to be binary. 
For simplicity, we use $\lambda_0 = \lambda_1 = \lambda_2 = \lambda_3 = 1.0$. \\[-4pt]

\par{\noindent \bf Discussion.\hspace{5pt}}
With the described training regime,
the model is forced to solve the proxy task by separately reconstructing two layers,
namely, the normal organ and tumor, 
and predicting a mask to indicate the assignment of each pixel.
During inference on real tumor data, 
such inferred mask thus becomes the tumor segmentation.
Note that, 
despite we only describe the training procedure on synthetic data, 
real tumor images can also be mixed into the training set,
{\em i.e.}~$x_{t_i} \in \mathcal{D}_{\text{tumor}}$, 
which simply requires to turn off the respective losses,
by setting $\lambda_0 = \lambda_1 = \lambda_2 = 0$,  $\alpha = 1$.
Experimentally, as shown in Table.~\ref{tab5}, 
this has been shown to further bridge the gap between simulation and real data, 
thus improving the performance of zero-shot tumor segmentation. 

\section{Experiments}
\label{sec4}

\subsection{Implementation Details}
Our proposed self-supervised learning method is flexible for any 3D encoder-decoder network.
For simplicity, we adopt the commonly used 3D UNet~\cite{iek20163DUL} as the network backbone,
which is same as the previous work~\cite{Zongwei2019}.
Our framework is trained with PyTorch on a GeForce RTX 3090 GPU with 24GB memory. 
At the {\em pre-training} stage, 
the volumetric input has a size of $128\times 128\times 128$ with the mini-batch size of 2. 
We adopt Adam optimiser~\cite{Kingma2014} with an initial learning rate of $10^{-3}$, 
and decayed with ReduceLROnPlateau scheduler. 
During {\em inference}, 
the predicted tumor mask can be treated as the result of zero-shot segmentation.
At the {\em fine-tuning} stage, 
the pre-trained encoder is used as an initialization for the network,
and a randomly initialized decoder with skip connections is added for supervised fine-tuning.
Note that all experiments during fine-tuning use data augmentations proposed in nnU-Net~\cite{Isensee2021},
unless otherwise noted.
The entire model is trained end-to-end with Dice loss, 
using Adam~\cite{Kingma2014} optimiser with an initial learning rate of $10^{-4}$.

\subsection{Pretraining Datasets}
\vspace{3pt}
\par{\noindent \bf Brain MRI Dataset.\hspace{3pt}}
We collected 993 healthy brain Fluid-Attenuated Inversion Recovery (FLAIR) MRI from OASIS dataset~\cite{Marsden_OASIS}, a longitudinal neuroimaging, clinical, 
and cognitive dataset for normal aging and Alzheimer Disease. 
All volumes are already co-registered, 
skull stripped and resampled in atlas space.\\[-5pt]

\par{\noindent \bf Abdomen CT Dataset.\hspace{3pt}}
We collected 663 abdomen CT scans from 6 public datasets, including BTCV~\cite{Landman_BTCV}, CHAOs~\cite{Kavur_2021}, KiTS~\cite{Nicholas_KiTS}, 
Pancreas-CT~\cite{Roth2015DeepOrganMD}, 
and two subsets ({\em i.e.} Pancreas and Spleen) from the Medical Segmentation Decathlon (MSD) challenge~\cite{simpson2019large}. 
We extracted the liver ROI of all the volumes with an off-the-shelf model based on nnU-Net~\cite{Isensee2021}, and our self-supervised training only uses these ROIs.

\subsection{Downstream Tasks and Benchmarks}
\vspace{3pt}
\par{\noindent \bf Brain Tumor Segmentation. \hspace{3pt}} 
BraTS2018~\cite{Menze2015} contains 285 training cases and 66 validation cases acquired by different MRI scanners. 
For each patient, T1 weighted, post-contrast T1-weighted, T2-weighted and FLAIR MRIs are provided. 
Each tumor is categorised into edema, necrosis and non-enhancing tumor and enhancing tumor. 
Each image is normalized independently by subtracting its mean and dividing by its standard deviation.
We conduct two experiments on this dataset: \\[-12pt]
\begin{itemize}[leftmargin=0.15in]
\setlength\itemsep{1pt}
\item {\bf WTS:} Whole tumor segmentation with FLAIR images. 
We adopt 3-fold cross-validation on the training cases, 
in which two folds (190 patients) are for training and one fold (95 patients) for test. 
\item {\bf BTS:} 
Whole Tumor (WT), Enhanced Tumor (WT) and Tumor Core (TC) segmentation with four modality images. 
Note that, the pre-training dataset for brain tumor segmentation contains data from only one modality, FLAIR. 
Following the official BraTS challenge, 
we report the results of per class dice, sensitivity, 
specificity and Hausdorff distances on the validation cases. 
Results are averages over three random trials. 
\end{itemize}

\vspace{3pt}
\par{\noindent \bf Liver Tumor Segmentation. \hspace{3pt}}
LiTS2017~\cite{bilic2019liver} contains 131 labeled CT scans acquired at seven hospitals and research centers, with each scan being paired with the liver and liver tumor boundaries traced manually by radiologists.  \\[-12pt]

\begin{itemize}[leftmargin=0.15in]
\item {\bf LTS:} Liver Tumor segmentation. 
We select 118 CT scans with tumor and extract the liver ROI based on the liver mask.
We adopt 3-fold cross-validation, in which two folds (79 patients) are for training and one fold (39 patients) for test. 
\end{itemize} 

\par{\noindent \bf Summary:} 
As for downstream tasks, we consider tumor segmentation in brain and liver, 
after self-supervised training models on two pretraining datasets respectively, 
and evaluate them with two different protocols. 
Specifically, WTS and LTS are used for evaluating the transferability and generalisability of models via supervised fine-tuning and zero-shot segmentation, 
while BTS is only used for supervised fine-tuning experiments,
since it includes fine-grained tumor categorisation. 

\subsection{Baseline Methods}
\label{sec4.2}
We compare against the state-of-the-art self-supervised learning approaches.
In general, these approaches can be cast into two categories,
namely, single-stage ({\em{i.e.}}~unsupervised anomaly detection: 
{\bf 3D IF}~\cite{marimont2021implicit}, 
{\bf ASC-Net}~\cite{Dey2021ASCNetA}) 
and two-stage ({\em{i.e.}}~pre-training and fine-tuning:
{\bf 3D RPL, 3D Jigsaw, 3D Rotation}~\cite{NEURIPS2020_d2dc6368} and
{\bf Model Genesis}~\cite{Zongwei2019}) methods. 
In essence, all self-supervised learning approaches differ only in the definition of the proxy tasks. See supplementary material for the details of the baseline methods.

\begin{table}[t]
\footnotesize
\centering
\setlength{\tabcolsep}{6pt}
\begin{tabular}[t]{c|ccc|cc}
\toprule
& \multicolumn{3}{c|}{Transformation} &  \multicolumn{2}{c}{Target Task~(Dice~$\%$ $\uparrow$)} \\[1.5pt]
Model &  Intensity & Blur & Elastic & WTS & LTS\\ \midrule
A0&  & & & 52.62 & 5.88\\[1pt]
A1& $\checkmark$ & & & 67.64 & 33.95 \\[1pt]
A2& & $\checkmark$ & & 62.28 & 10.1 \\[1pt]
A3& & & $\checkmark$ & 59.91 & 3.47 \\[1pt]
A4&$\checkmark$ & $\checkmark$ & & 69.53 & 38.91 \\[1pt]
A5&$\checkmark$ & & $\checkmark$ & 70.97 & 36.15 \\[1pt]
A6&$\checkmark$ & $\checkmark$ & $\checkmark$ & \textbf{71.63} & \textbf{40.78}\\
\midrule
\multicolumn{4}{c}{Real tumor (Upper Bound)} & 72.83 & 42.58\\
\bottomrule
\end{tabular}
\caption{Ablation study on data simulation. 
Results are reported for zero-shot tumor segmentation.
}
\label{tab1}
\end{table}

\section{Results}
\label{sec5}
In this section, we report the experimental results.
To begin with, 
we conduct extensive ablation studies,
as shown in Section~\ref{sec5.1},
to investigate the impact of different hyper-parameters during artificial ``tumor'' generation process 
and the necessity of proxy tasks.

Next, in Section~\ref{sec5.2}, 
we compare with the state-of-the-art unsupervised anomaly detection approaches for different datasets on {\em zero-shot} tumor segmentation.
When some manual annotations are available.
In Section~\ref{sec5.3}, 
we evaluate the model's transferability by fine-tuning the model with $0\%$ to $100\%$ segmentation masks.
Aligning with previous works,
we treat model trained from self-supervised learning as an initialization for supervised fine-tuning. 
Lastly, we fine-tune the self-supervised model end-to-end on downstream tasks with full annotation in Section~\ref{sec5.4}.

\subsection{Ablation Studies}
\label{sec5.1}
We conduct ablation studies to understand the effect of different choices on artificial transformations in data simulation and the proxy tasks.
Specifically, 
we train multiple self-supervised models by varying one factor at a time, 
and report zero-shot segmentation results. \\[-4pt]

{\noindent \bf On Data Simulation. \hspace{3pt}}
To understand the quality of our simulation procedure, 
we estimate a performance upper bound for our proposed simulation process, 
with an Oracle test.
Specifically,
instead of artificially generating tumors and textures,
we take the ground-truth masks, to crop the tumor and blend it into the normal image,
and train our self-supervised model on such generated data.

As shown in Table~\ref{tab1}, we make the following observations: 
(1) Compared to the case where no transformation is adopted, 
{\em i.e.}~directly blend one region cropped from another image, 
artificial transformations always bring significant improvements,
demonstrating the necessity of designing suitable transformations;
(2) Linear transformation acts as the basis for all the transformations. 
This is consistent with the fact that intensity contrast has always been 
an effective cue to differentiate lesions from normal organs;
(3) Our simulated tumor shows comparable results with the estimated upper bound,
indicating that, with sufficient texture randomisation,
models trained on simulated data can indeed generalise to segmenting real tumors. \\[-4pt]

\begin{table}[t]
\footnotesize
\centering
\setlength{\tabcolsep}{6pt}
\begin{tabular}[t]{c|cccc|cc}
\toprule
& \multicolumn{4}{c|}{Optimisation} &  \multicolumn{2}{c}{Target Task~(Dice $\%$ $\uparrow$)} \\[2pt]
Model & $\mathcal{L}_0$ & $\mathcal{L}_1$ & $\mathcal{L}_2$ & $\mathcal{L}_3$ & WTS & LTS \\
\midrule
B0 & \checkmark & & & & 38.47 & 7.24 \\[1pt]
B1 & & & \checkmark & & 66.84 & 37.09 \\[1pt]
B2 & \checkmark & \checkmark & & & 64.66 & 33.08\\[1pt]
B3 & \checkmark & \checkmark & \checkmark & & 70.92 & 40.07 \\[1pt]
B4 &  \checkmark & \checkmark & \checkmark & \checkmark & \textbf{71.63} & \textbf{40.78}\\
\midrule
\multicolumn{5}{c}{Ours~(w real tumor data)} & 72.62 & 41.08\\
\bottomrule
\end{tabular}
\vspace{-.2cm}
\caption{Ablation study on proxy tasks. 
Results are reported for zero-shot tumor segmentation.
\vspace{-.2cm}
}
\label{tab2}
\end{table}

\begin{figure*}[!htb]
\hspace{-2pt}
\centering{\includegraphics[width=\textwidth]{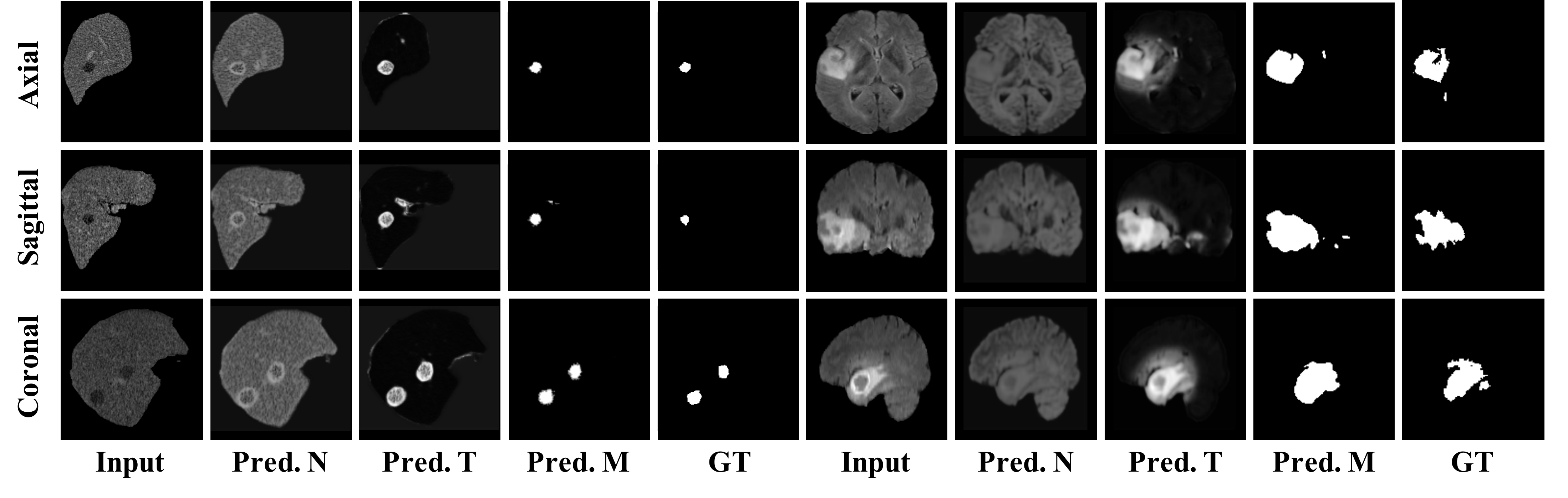}}
\vspace{-.8cm}
\caption{Zero-shot tumor segmentation from our self-supervised model on real tumor data.
The rows from top to bottom present the slices viewed from axial, coronal, and sagittal views. 
For liver tumor (left) and brain tumor (right), we show: 
Input, Predicted normal image (Pred.N), Predicted tumor image (Pred.T), Predicted mask (Pred.M), Ground Truth (GT).
\vspace{-.2cm}
}
\label{fig4}
\end{figure*}

{\vspace{0.1cm} \noindent \bf On Proxy Tasks. \hspace{2pt}}
Here, 
we conduct ablation studies to verify the usefulness of different losses in Eq.~\eqref{eq10}.
As shown in Table~\ref{tab2},
we first summarise the different combinations of proxy task as follows:
(1) $\mathcal{L}_0$, 
refers to a restoration task that recovers the normal organ from tumor input, 
and segmentation can be obtained by thresholding the reconstruction error;
(2) $\mathcal{L}_2$, 
refers to the proxy task for segmenting synthetic anomalies with discriminative training;
(3) $\mathcal{L}_0 + \mathcal{L}_1$, 
learns to distinguish the abnormal region by reconstructing both the normal and tumor parts;
(4) $\mathcal{L}_0 + \mathcal{L}_1 + \mathcal{L}_2$, 
refers to training without restoring input volume;
(5) $\mathcal{L}_0 + \mathcal{L}_1 + \mathcal{L}_2 + \mathcal{L}_3$, 
refers to the full self-supervised learning approach we proposed;
(6) {\em Ours (w real tumor data)}, denotes the case, 
where real tumor data is also introduced to the training process,
though only the full volume reconstruction~($\mathcal{L}_3$) is adopted.

From the experimental results, 
we can observe that the segmentation from restoration~($\mathcal{L}_0$) is far from being satisfactory, 
this is not surprising as such proxy task is not specifically designed for the tumor segmentation task.
In contrast, 
learning to segment synthetic anomalies with a discriminative loss~($\mathcal{L}_2$) substantially improves the segmentation results,
and combining all proxy tasks give the best performance.
Besides, when mixing real tumor data for self-supervised training,
despite only the volume reconstruction~($\mathcal{L}_3$) is available,
it is still beneficial to bridge the domain gap between simulated and real data, thus, enhance the zero-shot segmentation.

\subsection{Comparison on Zero-shot Tumor Segmentation}
\label{sec5.2}
In this section, 
we compare with state-of-the-art unsupervised anomaly detection methods.
As shown in Table \ref{tab3}, our approach surpasses the other ones by a large margin,
in particular, on brain tumor segmentation~(WTS), 
we substantially improve the Dice score over the recent state-of-the-art methods from $67.20\%$ to $71.63\%$, and from $32.24\%$ to $40.78\%$ on liver tumor segmentation~(LTS). 
In Figure~\ref{fig4},
we show qualitative results from our self-supervised model on real tumor data under a zero-shot setting, 
{\em i.e.}~models are directly adopted for mask prediction {\em without} using any manual annotation whatsoever.
As can be seen,
our method accurately detects and segments the tumors,
along with the intensity distribution of the restored image closely resembling that of the normal organ,
effectively inpainting the tumor regions.
However, note that we do not favor high-quality reconstructions,
what we really care about is the tumor segmentation mask.
More visualisation can be found in the supplementary material.

\subsection{Analysis on Model Transferability}
\label{sec5.3}
In addition to evaluating on zero-shot segmentation, 
we also study the transferability of proposed self-supervised learning method by finetuning on different ratios of groundtruth masks, and evaluating on the same test set.
To compare with the existing methods on the same pre-training datasets,
we adopt the official code released by the authors, and re-train it.
As shown in Table~\ref{tab4}, we conduct experiments on WTS and LTS,
and report Dice score~(most widely considered metric in segmentation).
For both tasks, our proposed method shows superior performance on all supervision levels.
Impressively, 
our self-supervised model really shines under a low-data regime, {\em e.g.}~$10\%$.
On zero-shot segmentation with $0\%$ annotation available,
unsurprisingly, 
our model clearly outperforms all other two-stage approaches, 
as their pre-training stage is unable to give segmentation predictions,
thus must rely on supervised finetuning.

\begin{table}[!t]
\footnotesize
\centering
\begin{threeparttable}
\setlength{\tabcolsep}{3pt}
\begin{tabular}{ccccc}
\toprule
\multirow{2}*{Method} & \multirow{2}*{Year} & \multirow{2}*{Supervision} & \multicolumn{2}{c}{Dice ($\%$) $\uparrow$ } \\
 \cmidrule(lr){4-5} & & & WTS & LTS \\
\toprule
Scratch~(w.~aug) & -- & $\checkmark$ & $86.03 \pm 0.90$ & $60.04 \pm 4.03$   \\
\midrule
3D IF~\cite{marimont2021implicit} & 2021 & \xmark & $67.20 \pm 15.5^*$ & -- \\[1pt]
ASC-Net~\cite{Dey2021ASCNetA} & 2021 & \xmark & $63.67^*$ & $32.24^*$ \\[1pt]
{\bf Ours} & 2021 & \xmark & {\bf 71.63 $\pm$ 0.84 } & {\bf 40.78 $\pm$ 0.43}\\
\bottomrule
\end{tabular}
\end{threeparttable}
\vspace{-.2cm}
\caption{Compare with SOTA unsupervised anomaly segmentation methods on whole tumor segmentation. Results with $*$ are directly copied from original papers~\cite{marimont2021implicit, Dey2021ASCNetA}.
\vspace{-.2cm}
}
\label{tab3}
\end{table}

\begin{table*}[!htb]
\scriptsize
\centering
\setlength{\tabcolsep}{7pt}
\begin{tabular}{c>{\columncolor{mygray}}ccccccc>{\columncolor{mygray}}ccccccc}
\toprule
\multirow{2}*{Method} & \multicolumn{7}{c}{WTS  (Dice$\%$ $\uparrow$)}  &  \multicolumn{7}{c}{LTS  (Dice$\%$ $\uparrow$)} \\
\cmidrule(lr){2-8} \cmidrule(lr){9-15}
& $0\%$ & $10\%$ & $20\%$ & $40\%$ & $60\%$ & $80\%$ & $100\%$ & $0\%$ & $10\%$ & $20\%$ & $40\%$ & $60\%$ & $80\%$ & $100\%$\\
\toprule
Scratch~(w.~aug) & 8.62 & 77.18 & 77.10 & 81.47 & 83.46 & 83.75 & 86.03 & 3.52 & 45.43 & 52.96 & 52.80 & 53.57 & 56.59 & 60.04\\[1pt]
3D RPL~\cite{NEURIPS2020_d2dc6368} & 6.79 & 78.85 & 79.00 & 81.68 & 82.63 & 84.29 & 85.64 & 3.11 & 49.25 & 49.51 & 55.53 & 56.93 & 57.60 & 60.05 \\[1pt]
Genesis~\cite{Zongwei2019} & 6.13 & 77.29 & 77.49 & 80.35 & 83.62 & 83.96 & 85.62 & 2.01 & 48.87 & 54.37 & 54.21 & 57.93 & 58.72 & 61.34\\[1pt]
{\bf Ours} & {\bf 71.63} & {\bf 79.80} & {\bf 79.66} & {\bf 82.64} & {\bf 83.92} & {\bf 84.49} & {\bf 86.13} & {\bf 40.78} & {\bf 53.61} & {\bf 54.84} & {\bf 57.74} & {\bf 59.36} & {\bf 60.52} & {\bf 61.91}\\
\bottomrule
\end{tabular}
\vspace{-.2cm}
\caption{Compare with SOTA self-supervised methods by fine-tuning on different ratio of labeled data. We report the Dice score.
Zero-shot tumor segmentation has been colored with gray.
With less labeled data, 
our method clearly outperforms all other two-stage approaches.
}
\label{tab4}
\end{table*}

\begin{table*}[!htb]
\scriptsize
\centering
\setlength{\tabcolsep}{5pt}
\begin{tabular}{ccccccccc}
\toprule
\multirow{2}*{Method} & \multicolumn{2}{c}{Dice ($\%$) $\uparrow$ } & \multicolumn{2}{c}{Hausdorff ($mm$) $\downarrow$} & \multicolumn{2}{c}{Sensitivity ($\%$) $\uparrow$} & \multicolumn{2}{c}{Specificity ($\%$) $\uparrow$ }\\
 \cmidrule(lr){2-3} \cmidrule(lr){4-5} \cmidrule(lr){6-7} \cmidrule(lr){8-9}
 &  WTS & LTS &  WTS & LTS  &  WTS & LTS  & WTS & LTS \\
\toprule
Scratch (w.o. aug) & $83.32 \pm 2.42$ & $58.30 \pm 2.93 $ & $9.84 \pm 1.16 $ & $40.15 \pm 3.61 $ & $80.49 \pm 2.56$ & $ 59.15 \pm 7.56$ & $99.89 \pm 0.02$& $99.64 \pm 1.45 $ \\[1pt]
Scratch (w. aug) & $86.03 \pm 0.90$ & $60.04 \pm 4.03$ & \textbf{7.45 $\pm$ 0.24} & $30.70 \pm 5.87 $ & $ 85.21 \pm 2.93 $ & $59.19 \pm 7.83 $ & $99.90 \pm 0.03 $ & \textbf{99.79 $\pm$ 1.04} \\ \midrule
3D RPL~\cite{NEURIPS2020_d2dc6368} & $85.64 \pm 0.34 $ & $ 60.15 \pm 4.36 $ & $8.60\pm 1.37 $ & $32.39 \pm 5.84 $ & $86.06 \pm 2.21 $ & $ 59.21 \pm 9.23 $ & $99.88 \pm 0.01 $ & $99.73 \pm 1.91 $ \\[1pt]
3D Jigsaw~\cite{NEURIPS2020_d2dc6368} & $84.95 \pm 0.97 $ & $ 59.13 \pm 5.12 $ & $9.01\pm 0.54 $ & \textbf{29.20 $\pm$ 3.66} & $83.20 \pm 7.84 $ & $59.54 \pm 9.90 $ & $99.90 \pm 0.02 $ & $99.68 \pm 2.72 $ \\[1pt]
3D Rotation~\cite{NEURIPS2020_d2dc6368} & $84.35 \pm 1.12 $ & $59.72\pm 4.87 $ & $10.98 \pm 2.65 $ & $33.33 \pm 16.56 $ & $84.21 \pm 2.25 $ & $ 59.91\pm 8.57 $ & $99.89 \pm 0.01$ & $99.86 \pm 0.05 $ \\[1pt]
Genesis~\cite{Zongwei2019} & $85.62 \pm 0.48$ & $61.34 \pm 4.26 $ & $8.01 \pm 0.77$ & $39.92 \pm 8.90$ & $84.56 \pm 1.30$ & \textbf{65.59 $\pm$ 10.11} & $99.90\pm 0.02$ & $99.73\pm 1.74$ \\[1pt]
{\bf Ours}  &\textbf{86.13 $\pm$ 0.91} & \textbf{61.91 $\pm$ 2.71} & $8.11 \pm 0.54$ & $31.41 \pm 8.33 $ & \textbf{85.49 $\pm$ 2.96} & $65.39 \pm 3.72$ & \textbf{99.90 $\pm$ 0.01} & $99.73 \pm 1.89$ \\
\bottomrule
\end{tabular}
\vspace{-.2cm}
\caption{Compare with SOTA self-supervised methods on WTS (brain whole tumor segmentation) and LTS (liver tumor segmentation) with fully supervised fine-tuning.
We report the results of mean $\pm$ std over 3-fold cross validation.
Scratch (w.o.aug) refers to model trained from scratch without using data augmentation.
}
\label{tab5}
\end{table*}

\begin{table*}[!htb]
\scriptsize
\centering
\setlength{\tabcolsep}{6pt}
\begin{tabular}{ccccccccccccc}
\toprule
\multirow{2}*{Method} & \multicolumn{3}{c}{Dice ($\%$) $\uparrow$ } & \multicolumn{3}{c}{Hausdorff ($mm$) $\downarrow$ } & \multicolumn{3}{c}{Sensitivity ($\%$) $\uparrow$ } & \multicolumn{3}{c}{Specificity ($\%$) $\uparrow$ }\\
 \cmidrule(lr){2-4} \cmidrule(lr){5-7} \cmidrule(lr){8-10} \cmidrule(lr){11-13}
 & WT & TC & ET & WT & TC & ET & WT & TC & ET & WT & TC & ET \\
\toprule
Rubik's Cube+~\cite{ZHU2020101746} & $89.60$ & $80.42$ & $75.10$ & $22.32$ & $13.88$ & $13.26$ & $88.53$ & $\bf{82.61}$ & $81.85$ & $\bf{99.70}$ & $99.84$ & $99.85 $\\
\midrule
Scratch & $84.23 $ & $ 64.44$ & $69.31$ & $ 19.94$ & $19.85$  & $10.98$ & $83.07$ & $71.89$ & $67.80$ & $99.35$  & $99.21$ & $99.85$ \\
(w.o.~aug) & ${\pm 0.95} $ & ${\pm 3.65}$ & ${\pm 5.08}$ & ${\pm 3.97}$ & ${\pm 1.03}$  & ${\pm1.83}$ & ${\pm 0.97}$ & ${\pm 3.44}$ & ${\pm 1.30}$ & ${\pm 1.60}$  & ${\pm 0.94}$ & ${\pm 0.01}$\\ [1.5pt]
Scratch & $89.70$ & $80.53$ & $77.15$ & $9.47$ & $8.50$ & $\bf 3.76$ & $\bf 90.35$ & $77.48$ & $79.60$ & $99.47$ & $\bf 99.87$ & $99.83$\\
(w.~aug) & ${\pm 0.27}$ & ${\pm 0.25}$ & ${\pm 0.31}$ & ${\pm 5.10}$ & ${\pm 1.88}$ & $\bf {\pm 1.38}$ & $\bf {\pm 0.10}$ & ${\pm 0.52}$ & ${\pm 0.34}$ & ${\pm 0.04}$ & $\bf {\pm 0.06}$ & ${\pm 0.01}$  \\ \midrule
\multirow{2}*{3D RPL~\cite{NEURIPS2020_d2dc6368}} & $88.40$ & $80.92$ & $77.04$ & $10.59$ & $8.96$ & $6.15$ & $87.71$ & $79.33$ & $80.57$ & $ 99.56$ & $99.84$ & $99.81$ \\
& ${\pm 1.64}$ & ${\pm 1.06}$ & ${\pm 0.44}$ & ${\pm 6.03}$ & ${\pm 0.69}$ & ${\pm 2.01}$ & ${\pm 2.98}$ & ${\pm 0.99}$ & ${\pm 1.68}$ & $ {\pm 0.06}$ & ${\pm 0.01}$ & ${\pm 0.02}$ \\ [1.5pt]
\multirow{2}*{3D Jigsaw~\cite{NEURIPS2020_d2dc6368}} & $89.33$ & $79.07$ & $76.12$ & $8.42$ & $ 9.67$ & $5.01$ & $89.84$ & $78.58$ & $79.76$ & $99.46$ & $99.80$ & $99.82$ \\
& ${\pm 0.32}$ & ${\pm 0.35}$ & ${\pm 0.19}$ & ${\pm 4.26}$ & ${\pm 1.98}$ & ${\pm 0.42}$ & ${\pm 0.46}$ & ${\pm 1.78}$  & ${\pm 0.60}$ & ${\pm 0.06}$ & ${\pm 0.06}$ & ${\pm 0.01}$\\ [1.5pt]
\multirow{2}*{3D Rotation~\cite{NEURIPS2020_d2dc6368}} & $89.22$ & $78.82$ & $74.92$ & $10.45$ & $10.67$ & $5.88$ & $89.72$ & $78.32$ & $78.94$ & $99.42$ & $99.81$ & $99.82$ \\
&${\pm 0.35}$ & ${\pm 0.27}$ & ${\pm 1.42}$ & ${\pm 1.33}$ & ${\pm 2.17}$ & ${\pm 1.73}$ & ${\pm 0.71}$ & ${\pm 2.21}$ & ${\pm 1.38}$ & ${\pm 0.08}$ & ${\pm 0.07}$ & ${\pm 0.01}$ \\ [1.5pt]
\multirow{2}*{Genesis~\cite{Zongwei2019}} & $\bf 90.10$ & $\bf 82.77$ & $75.78$ & $6.36$ & $\bf 7.63$ & $4.68$ & $89.96$ & $ 82.52$ & $\bf 82.23$ & $99.54$ & $99.80$ & $99.80$ \\
& $\bf{\pm 0.64}$ & $\bf {\pm 0.26}$ & ${\pm 0.36}$ & ${\pm 1.21}$ & ${\pm 1.04}$ & ${\pm 1.24}$ & ${\pm 1.51}$ & $\bf {\pm 2.04}$ & $\bf {\pm 2.16}$ & ${\pm 0.06}$ & ${\pm 0.05}$ & ${\pm 0.02}$ \\ [1.5pt]
\multirow{2}*{{\bf Ours}} &  $\bf 90.10$ & $82.47$ & $\bf 77.32$ & $\bf 5.89$ & $7.78$ & $4.38$ & $90.05$ & $80.01$ & $81.39$ & $99.54$ & $99.80$ & $\bf 99.86$\\
& $\bf{\pm 0.51}$ & ${\pm 0.84}$ & $\bf {\pm 0.06}$ & $\bf {\pm 0.40}$ & ${\pm 0.75}$ & ${\pm 0.60}$ & ${\pm 0.30}$ & ${\pm 0.55}$ & ${\pm 0.64}$ & ${\pm 0.01}$ &${\pm 0.01}$ & $\bf {\pm 0.01}$ \\ \bottomrule
\end{tabular}
\vspace{-.2cm}
\caption{Compare with SOTA self-supervised methods on BTS~(whole tumor,
tumor core and enhanced tumor segmentation) with fully supervised fine-tuning.
We report the results of mean $\pm$ std by fixed the training set and run three different seeds.
The results of Rubik's Cube+ are directly copied from the paper~\cite{ZHU2020101746}.
Scratch (w.o.aug) refers to model trained from scratch without using data augmentation.
}
\label{tab6}
\end{table*}

\subsection{Fully Supervised Fine-tuning}
\label{sec5.4}
This section compares with other self-supervised approaches under the two-stage evaluation procedure, where the pre-trained models are finetuned with all groundtruth masks.
As shown in Table~\ref{tab5} and Table~\ref{tab6},
we report the Dice, Hausdorff, Sensitivity and Specificity for all three tasks, 
{\em i.e.}~WTS, LTS, and BTS.
We observe two phenomena:
{\em First}, all self-supervised approaches clearly demonstrate superior performance over the scratch model without augmentation, 
however, such benefits quickly diminish once we introduce more data augmentations.
We conjecture this is because pretraining and finetuning are using dataset of similar scale,
whereas for self-supervised to be useful, 
the amount of unlabelled data has to be significantly larger.
This is consistent with the results reported in previous work~\cite{NEURIPS2020_d2dc6368}. 
{\em Second}, none of the existing approaches dominate all metrics, 
and in fact, the variance in performance between different approaches is typically less than $2\%$. Nevertheless, most of our model's performance are among the top two with lower standard deviations, which implies the robustness of our proposed self-supervised training regime.  

\subsection{Limitations}
\label{sec5.5}
Here, we briefly discuss some of the limitations in this work:
First, 
there remains domain gap between the synthetic and the real tumor data,
which can be observed from the fact that zero-shot segmentation still underperforms the supervised finetuning;
Second,
the model at the moment is not able to do fine-grained tumor categorisation,
since it remains unclear how to simulate textures for the inner structure of a tumor. We speculate that such issue could be alleviated by further introducing GAN~\cite{Goodfellow2014GenerativeAN} or VQ-VAE~\cite{Oord2017NeuralDR} into the framework, which we leave for future work.

\section{Conclusion}
To conclude, in this paper, 
we propose a novel self-supervised learning method for 3D tumor segmentation, 
exploiting synthetic data to train a ``layer decomposition'' model. 
The key difference to the conventional self-supervised learning model is that, 
our model can be directly applied to the downstream tumor segmentation task,
advocating the zero-shot tumor segmentation.
To validate the effectiveness of our approach, 
we conducted extensive experiments on two benchmarks: 
BraTS2018 for brain tumor segmentation and LiTS2017 for liver tumor segmentation,
and show superior performance on transferability and generalisability to strong baselines over the state-of-the-art methods.


\bibliographystyle{ieee_fullname}
\bibliography{ref}

\begin{thebibliography}{10}\itemsep=-1pt

\bibitem{Parvez2021}
Parvez Ahmad, Saqib Qamar, Linlin Shen, and Adnan Saeed.
\newblock Context aware 3d unet for brain tumor segmentation.
\newblock In {\em Brainlesion: Glioma, Multiple Sclerosis, Stroke and Traumatic
  Brain Injuries}, pages 207--218, 2021.

\bibitem{alayrac2019visual}
Jean-Baptiste Alayrac, Joao Carreira, and Andrew Zisserman.
\newblock The visual centrifuge: Model-free layered video representations.
\newblock In {\em Proceedings of the IEEE Conference on Computer Vision and
  Pattern Recognition}, pages 2457--2466, 2019.

\bibitem{Bai2019SelfSupervisedLF}
Wenjia Bai, Chen Chen, G. Tarroni, J. Duan, Florian Guitton, S. Petersen, Yike
  Guo, P. Matthews, and D. Rueckert.
\newblock Self-supervised learning for cardiac mr image segmentation by
  anatomical position prediction.
\newblock In {\em International Conference on Medical Image Computing and
  Computer Assisted Intervention}, pages 541--549, 2019.

\bibitem{BAUR2021101952}
Christoph Baur, Stefan Denner, Benedikt Wiestler, Nassir Navab, and Shadi
  Albarqouni.
\newblock Autoencoders for unsupervised anomaly segmentation in brain mr
  images: A comparative study.
\newblock {\em Medical Image Analysis}, 69:101952, 2021.

\bibitem{bilic2019liver}
Patrick Bilic, Patrick~Ferdinand Christ, Eugene Vorontsov, et~al.
\newblock The liver tumor segmentation benchmark (lits), 2019.

\bibitem{Brostow791190}
Gabriel Brostow and Irfan Essa.
\newblock Motion based decompositing of video.
\newblock In {\em Proceedings of IEEE International Conference on Computer
  Vision}, 1999.

\bibitem{Sintel2012}
Daniel~J. Butler, Jonas Wulff, Garrett~B. Stanley, and Michael~J. Black.
\newblock A naturalistic open source movie for optical flow evaluation.
\newblock In {\em Proceedings of the European Conference on Computer Vision},
  pages 611--625, 2012.

\bibitem{Caron_2018_ECCV}
Mathilde Caron, Piotr Bojanowski, Armand Joulin, and Matthijs Douze.
\newblock Deep clustering for unsupervised learning of visual features.
\newblock In {\em Proceedings of the European Conference on Computer Vision},
  2018.

\bibitem{Chen2019SelfSupervisedLF}
Liang Chen, Paul Bentley, Kensaku Mori, Kazunari Misawa, Michitaka Fujiwara,
  and Daniel Rueckert.
\newblock Self-supervised learning for medical image analysis using image
  context restoration.
\newblock {\em Medical Image Analysis}, 58:101539, 2019.

\bibitem{Shuai2019}
Shuai Chen, Gerda Bortsova, Antonio Garc{\'i}a-Uceda~Ju{\'a}rez, Gijs van
  Tulder, and Marleen de Bruijne.
\newblock Multi-task attention-based semi-supervised learning for medical image
  segmentation.
\newblock In {\em International Conference on Medical Image Computing and
  Computer Assisted Intervention}, pages 457--465, 2019.

\bibitem{Chen2020ASF}
Ting Chen, Simon Kornblith, Mohammad Norouzi, and Geoffrey~E. Hinton.
\newblock A simple framework for contrastive learning of visual
  representations.
\newblock {\em ArXiv}, abs/2002.05709, 2020.

\bibitem{CHEPLYGINA2019280}
Veronika Cheplygina, Marleen {de Bruijne}, and Josien~P.W. Pluim.
\newblock Not-so-supervised: A survey of semi-supervised, multi-instance, and
  transfer learning in medical image analysis.
\newblock {\em Medical Image Analysis}, 54:280--296, 2019.

\bibitem{Dey2021ASCNetA}
Raunak Dey and Yi Hong.
\newblock Asc-net : Adversarial-based selective network for unsupervised
  anomaly segmentation.
\newblock In {\em International Conference on Medical Image Computing and
  Computer Assisted Intervention}, 2021.

\bibitem{doersch2015unsupervised}
Carl Doersch, Abhinav Gupta, and Alexei~A. Efros.
\newblock Unsupervised visual representation learning by context prediction.
\newblock In {\em Proceedings of the IEEE International Conference on Computer
  Vision}, 2015.

\bibitem{Doersch2019Sim2realTL}
Carl Doersch and Andrew Zisserman.
\newblock Sim2real transfer learning for 3d pose estimation: motion to the
  rescue.
\newblock In {\em Advances in Neural Information Processing Systems}, pages
  12929--12941, 2019.

\bibitem{Marsden_OASIS}
Marcus DS, Fotenos AF, Csernansky JG, Morris JC, and Buckner RL.
\newblock Open access series of imaging studies: longitudinal mri data in
  nondemented and demented older adults.
\newblock {\em Journal of Cognitive Neuroscience}, 22:2677--84, 2018.

\bibitem{Huang9311201}
Ye Fei, Chaoqin Huang, Cao Jinkun, Maosen Li, Ya Zhang, and Cewu Lu.
\newblock Attribute restoration framework for anomaly detection.
\newblock {\em IEEE Transactions on Multimedia}, pages 1--1, 2020.

\bibitem{Feng8953870}
Zeyu Feng, Chang Xu, and Dacheng Tao.
\newblock Self-supervised representation learning by rotation feature
  decoupling.
\newblock In {\em Proceedings of the IEEE Conference on Computer Vision and
  Pattern Recognition}, 2019.

\bibitem{DoubleDIP}
Yossi Gandelsman, Assaf Shocher, and Michal Irani.
\newblock "double-dip": Unsupervised image decomposition via coupled
  deep-image-priors.
\newblock In {\em Proceedings of the IEEE Conference on Computer Vision and
  Pattern Recognition}, 2019.

\bibitem{Gidaris2018iclr}
Spyros Gidaris, Praveer Singh, and Nikos Komodakis.
\newblock Unsupervised representation learning by predicting image rotations.
\newblock In {\em International Conference on Learning Representations}, 2018.

\bibitem{Goodfellow2014GenerativeAN}
Ian~J. Goodfellow, Jean Pouget-Abadie, Mehdi Mirza, Bing Xu, David
  Warde-Farley, Sherjil Ozair, Aaron~C. Courville, and Yoshua Bengio.
\newblock Generative adversarial nets.
\newblock In {\em Advances in Neural Information Processing Systems}, 2014.

\bibitem{Gupta2016synthetic}
Ankush Gupta, Andrea Vedaldi, and Andrew Zisserman.
\newblock Synthetic data for text localisation in natural images.
\newblock In {\em Proceedings of the IEEE Conference on Computer Vision and
  Pattern Recognition}, 2016.

\bibitem{Kaiming9157636}
Kaiming He, Haoqi Fan, Yuxin Wu, Saining Xie, and Ross Girshick.
\newblock Momentum contrast for unsupervised visual representation learning.
\newblock In {\em Proceedings of the IEEE Conference on Computer Vision and
  Pattern Recognition}, pages 9726--9735, 2020.

\bibitem{Nicholas_KiTS}
Nicholas Heller, Sean McSweeney, Matthew~Thomas Peterson, et~al.
\newblock An international challenge to use artificial intelligence to define
  the state-of-the-art in kidney and kidney tumor segmentation in ct imaging.
\newblock {\em Journal of Clinical Oncology}, 38:626--626, 2020.

\bibitem{hjelm2019learning}
R~Devon Hjelm, Alex Fedorov, Samuel Lavoie-Marchildon, Karan Grewal, Phil
  Bachman, Adam Trischler, and Yoshua Bengio.
\newblock Learning deep representations by mutual information estimation and
  maximization.
\newblock In {\em International Conference on Learning Representations}, 2019.

\bibitem{Isensee2021}
Fabian Isensee, Paul~F. Jaeger, Simon A.~A. Kohl, Jens Petersen, and Klaus~H.
  Maier-Hein.
\newblock nnu-net: a self-configuring method for deep learning-based biomedical
  image segmentation.
\newblock {\em Nature Methods}, 18:203--211, 2021.

\bibitem{ISIN2016317}
Ali Işın, Cem Direkoğlu, and Melike Şah.
\newblock Review of mri-based brain tumor image segmentation using deep
  learning methods.
\newblock {\em Procedia Computer Science}, 102:317--324, 2016.

\bibitem{Jaderberg2014SyntheticDA}
Max Jaderberg, Karen Simonyan, Andrea Vedaldi, and Andrew Zisserman.
\newblock Synthetic data and artificial neural networks for natural scene text
  recognition.
\newblock In {\em NeurIPS Deep Learning Workshop}, 2014.

\bibitem{Kavur_2021}
A.~Emre Kavur, N.~Sinem Gezer, Mustafa Barış, Sinem Aslan, et~al.
\newblock Chaos challenge - combined (ct-mr) healthy abdominal organ
  segmentation.
\newblock {\em Medical Image Analysis}, 69:101950, 2021.

\bibitem{ke2021bcnet}
Lei Ke, Yu-Wing Tai, and Chi-Keung Tang.
\newblock Deep occlusion-aware instance segmentation with overlapping bilayers.
\newblock In {\em Proceedings of the IEEE Conference on Computer Vision and
  Pattern Recognition}, 2021.

\bibitem{Kingma2014}
Diederik Kingma and Jimmy Ba.
\newblock Adam: A method for stochastic optimization.
\newblock In {\em ICLR}, volume~42, 2014.

\bibitem{Kumar1541236}
M.~Pawan Kumar, Philip H.~S. Torr, and Andrew Zisserman.
\newblock Learning layered motion segmentations of video.
\newblock In {\em Proceedings of IEEE International Conference on Computer
  Vision}, 2005.

\bibitem{Landman_BTCV}
Bennett Landman, Zhoubing Xu, Juan~Eugenio Igelsias, Martin Styner,
  Thomas~Robin Langerak, and Arno Klein.
\newblock Multi-atlas labeling beyond the cranial vault – workshop and
  challenge.
\newblock {\em Journal of Clinical Oncology}, 2015.

\bibitem{Xiaomeng2018}
Xiaomeng Li, Hao Chen, Xiaojuan Qi, Qi Dou, Chi-Wing Fu, and Pheng-Ann Heng.
\newblock H-denseunet: Hybrid densely connected unet for liver and tumor
  segmentation from ct volumes.
\newblock {\em IEEE Transactions on Medical Imaging}, 37(12):2663--2674, 2018.

\bibitem{lu2020}
Erika Lu, Forrester Cole, Tali Dekel, Weidi Xie, Andrew Zisserman, David
  Salesin, William~T Freeman, and Michael Rubinstein.
\newblock Layered neural rendering for retiming people in video.
\newblock In {\em SIGGRAPH Asia}, 2020.

\bibitem{Lu_2021_CVPR}
Erika Lu, Forrester Cole, Tali Dekel, Andrew Zisserman, William~T. Freeman, and
  Michael Rubinstein.
\newblock Omnimatte: Associating objects and their effects in video.
\newblock In {\em Proceedings of the IEEE Conference on Computer Vision and
  Pattern Recognition}, pages 4507--4515, 2021.

\bibitem{marimont2021implicit}
Sergio~Naval Marimont and Giacomo Tarroni.
\newblock Implicit field learning for unsupervised anomaly detection in medical
  images.
\newblock In {\em International Conference on Medical Image Computing and
  Computer Assisted Intervention}, 2021.

\bibitem{Mayer2016}
Nikolaus Mayer, Eddy Ilg, Philip Hausser, Philipp Fischer, Daniel Cremers,
  Alexey Dosovitskiy, and Thomas Brox.
\newblock A large dataset to train convolutional networks for disparity,
  optical flow, and scene flow estimation.
\newblock In {\em Proceedings of the IEEE Conference on Computer Vision and
  Pattern Recognition}, 2016.

\bibitem{Menze2015}
B.~H. {Menze} et~al.
\newblock The multimodal brain tumor image segmentation benchmark (brats).
\newblock {\em IEEE Transactions on Medical Imaging}, 34(10):1993--2024, 2015.

\bibitem{Shaobo2019}
Shaobo Min, Xuejin Chen, Zheng-Jun Zha, and Yongdong Zhang.
\newblock A two-stream mutual attention network for semi-supervised biomedical
  segmentation with noisy labels.
\newblock {\em AAAI Conference on Artificial Intelligence}, 33:4578--4585,
  2019.

\bibitem{noroozi2017unsupervised}
Mehdi Noroozi and Paolo Favaro.
\newblock Unsupervised learning of visual representations by solving jigsaw
  puzzles.
\newblock In {\em Proceedings of the European Conference on Computer Vision},
  2016.

\bibitem{WHO2020}
World~Health Organization.
\newblock International agency for research on cancer.
\newblock 2020.

\bibitem{pathak2016context}
Deepak Pathak, Philipp Krahenbuhl, Jeff Donahue, Trevor Darrell, and Alexei~A.
  Efros.
\newblock Context encoders: Feature learning by inpainting.
\newblock In {\em Proceedings of the IEEE Conference on Computer Vision and
  Pattern Recognition}, 2016.

\bibitem{random2018Peng}
Xue~Bin Peng, Marcin Andrychowicz, Wojciech Zaremba, and Pieter Abbeel.
\newblock Sim-to-real transfer of robotic control with dynamics randomization.
\newblock {\em 2018 IEEE International Conference on Robotics and Automation},
  2018.

\bibitem{Perez2021self}
Fernando P{\'e}rez-Garc{\'\i}a, Reuben Dorent, Michele Rizzi, et~al.
\newblock A self-supervised learning strategy for postoperative brain cavity
  segmentation simulating resections.
\newblock {\em International Journal of Computer Assisted Radiology and
  Surgery}, pages 1--9, 2021.

\bibitem{Roth2015DeepOrganMD}
Holger~R. Roth, Le Lu, and et al.
\newblock Deeporgan: Multi-level deep convolutional networks for automated
  pancreas segmentation.
\newblock In {\em International Conference on Medical Image Computing and
  Computer Assisted Intervention}, pages 556--564, 2015.

\bibitem{simpson2019large}
Amber~L. Simpson, Michela Antonelli, Spyridon Bakas, and et al.
\newblock A large annotated medical image dataset for the development and
  evaluation of segmentation algorithms, 2019.

\bibitem{Srinivasan2019}
Pratul Srinivasan, Richard Tucker, Jonathan Barron, Ravi Ramamoorthi, Ren Ng,
  and Noah Snavely.
\newblock Pushing the boundaries of view extrapolation with multiplane images.
\newblock In {\em Proceedings of the IEEE Conference on Computer Vision and
  Pattern Recognition}, pages 175--184, 2019.

\bibitem{NEURIPS2020_d2dc6368}
Aiham Taleb, Winfried Loetzsch, Noel Danz, Julius Severin, Thomas Gaertner,
  Benjamin Bergner, and Christoph Lippert.
\newblock 3d self-supervised methods for medical imaging.
\newblock In {\em Advances in Neural Information Processing Systems},
  volume~33, pages 18158--18172, 2020.

\bibitem{Tan2021DetectingOW}
Jeremy Tan, Benjamin Hou, T.~G. Day, John Simpson, D. Rueckert, and Bernhard
  Kainz.
\newblock Detecting outliers with poisson image interpolation.
\newblock In {\em International Conference on Medical Image Computing and
  Computer Assisted Intervention}, 2021.

\bibitem{to2021selfsupervised}
Minh-Son To, Ian~G Sarno, Chee Chong, Mark Jenkinson, and Gustavo Carneiro.
\newblock Self-supervised lesion change detection and localisation in
  longitudinal multiple sclerosis brain imaging.
\newblock In {\em International Conference on Medical Image Computing and
  Computer Assisted Intervention}, 2021.

\bibitem{tobin2017domain}
Josh Tobin, Rachel Fong, Alex Ray, Jonas Schneider, Wojciech Zaremba, and
  Pieter Abbeel.
\newblock Domain randomization for transferring deep neural networks from
  simulation to the real world.
\newblock In {\em IEEE/RSJ International Conference on Intelligent Robots and
  Systems}, 2017.

\bibitem{lsiTulsiani18}
Shubham Tulsiani, Richard Tucker, and Noah Snavely.
\newblock Layer-structured 3d scene inference via view synthesis.
\newblock In {\em Proceedings of the European Conference on Computer Vision},
  2018.

\bibitem{Aaron2019cpc}
Aaron van~den Oord, Yazhe Li, and Oriol Vinyals.
\newblock Representation learning with contrastive predictive coding, 2019.

\bibitem{Oord2017NeuralDR}
A{\"a}ron van~den Oord, Oriol Vinyals, and Koray Kavukcuoglu.
\newblock Neural discrete representation learning.
\newblock In {\em Advances in Neural Information Processing Systems}, 2017.

\bibitem{Adelson341105}
John Y.~A. Wang and Edward Adelson.
\newblock Layered representation for motion analysis.
\newblock In {\em Proceedings of the IEEE Conference on Computer Vision and
  Pattern Recognition}, pages 361--366, 1993.

\bibitem{Wang334981}
John Y.~A. Wang and Edward Adelson.
\newblock Representing moving images with layers.
\newblock {\em IEEE Transactions on Image Processing}, 3(5):625--638, 1994.

\bibitem{Yang_2021_ICCV}
Charig Yang, Hala Lamdouar, Erika Lu, Andrew Zisserman, and Weidi Xie.
\newblock Self-supervised video object segmentation by motion grouping.
\newblock In {\em Proceedings of IEEE International Conference on Computer
  Vision}, pages 7177--7188, 2021.

\bibitem{Yeung2021}
Pak~Hei Yeung, Ana Namburete, and Weidi Xie.
\newblock Sli2vol: Annotate a 3d volume from a single slice with
  self-supervised learning.
\newblock In {\em International Conference on Medical Image Computing and
  Computer Assisted Intervention}, 2021.

\bibitem{Zhang2021sar}
Xiaoman Zhang, Shixiang Feng, Yuhang Zhou, Ya Zhang, and Yanfeng Wang.
\newblock Sar: Scale-aware restoration learning for 3d tumor segmentation.
\newblock In {\em International Conference on Medical Image Computing and
  Computer Assisted Intervention}, 2021.

\bibitem{Zongwei2019}
Zongwei Zhou, Vatsal Sodha, Md~Mahfuzur~Rahman Siddiquee, Ruibin Feng, Nima
  Tajbakhsh, Michael~B. Gotway, and Jianming Liang.
\newblock Models genesis: Generic autodidactic models for 3d medical image
  analysis.
\newblock In {\em International Conference on Medical Image Computing and
  Computer Assisted Intervention}, pages 384--393, 2019.

\bibitem{ZHU2020101746}
Jiuwen Zhu, Yuexiang Li, Yifan Hu, Kai Ma, S.~Kevin Zhou, and Yefeng Zheng.
\newblock Rubik’s cube+: A self-supervised feature learning framework for 3d
  medical image analysis.
\newblock {\em Medical Image Analysis}, 64:101746, 2020.

\bibitem{Zhuang2019SelfSupervisedFL}
Xinrui Zhuang, Yuexiang Li, Yifan Hu, Kai Ma, Yujiu Yang, and Yefeng Zheng.
\newblock Self-supervised feature learning for 3d medical images by playing a
  rubik's cube.
\newblock In {\em International Conference on Medical Image Computing and
  Computer Assisted Intervention}, pages 420--428, 2019.

\bibitem{Zong2018DeepAG}
Bo Zong, Qi Song, Martin~Renqiang Min, Wei Cheng, C. Lumezanu, Dae ki Cho, and
  Haifeng Chen.
\newblock Deep autoencoding gaussian mixture model for unsupervised anomaly
  detection.
\newblock In {\em ICLR}, 2018.

\bibitem{iek20163DUL}
{\"O}zg{\"u}n Çiçek, Ahmed Abdulkadir, Soeren~S. Lienkamp, Thomas Brox, and
  Olaf Ronneberger.
\newblock 3d u-net: Learning dense volumetric segmentation from sparse
  annotation.
\newblock In {\em International Conference on Medical Image Computing and
  Computer Assisted Intervention}, pages 424--432, 2016.

\end{thebibliography}

\clearpage
\onecolumn
\appendix
\tableofcontents
\newpage

\section{Implementation Details for Data Simulation}
\subsection{Texture Simulation}
\begin{figure}[h]
\centering{\includegraphics[width=0.9\columnwidth]{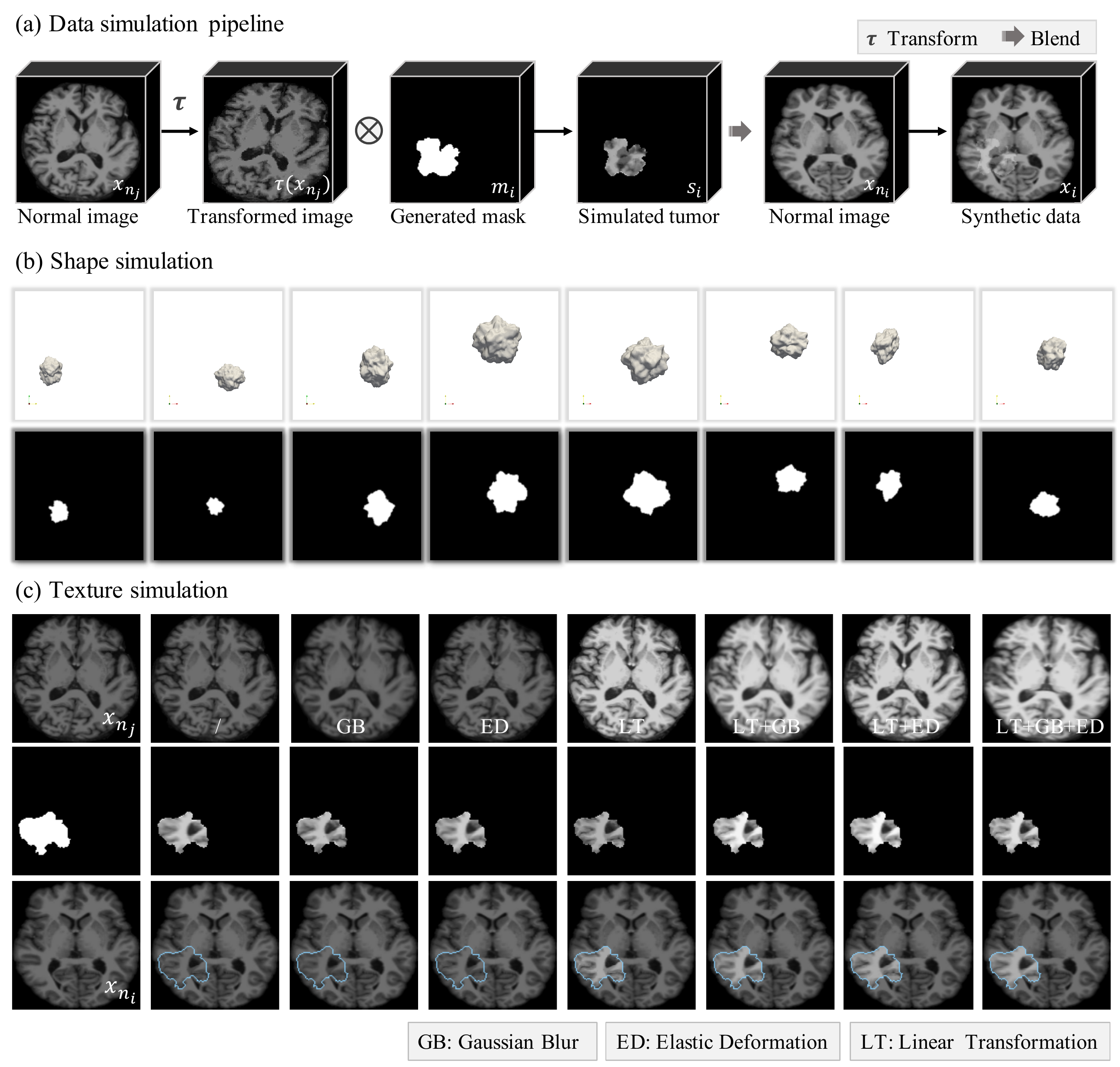}}
\caption{
\textbf{Illustration of the data simulation pipeline.} 
(a) Data simulation pipeline. 
We first simulate the tumor with a transformed image and a generated 3D mask, 
which provide texture and shape respectively. 
Then the synthetic data is composed by blending the simulated tumor into the normal image.
(b) Visualisation of shape simulation.
The first row presents the 3D polyhedrons with various shapes, sizes and locations,
and the second row shows the binary masks.
(c) Visualisation of texture simulation.
We have displayed a randomly selected normal image $x_{n_j}$ undergoing different transformations functions (top row),
the corresponding simulated tumor (second row),
and synthetic training data (bottom row) in columns 2$-$8. 
} 
\label{supple_fig_1}
\end{figure}

We adopt three commonly-used transformations: linear transformation, 
elastic deformation and Gaussian blur. 
Figure~\ref{supple_fig_1} shows the visualisation of the proposed image transformations used for texture simulation. 
In practise, we combine these three transformations, 
like $LT+GB+ED$, which can be formulated as:
\begin{align}
    \tau(x_{n_j}) = \tau_{LT}(\tau_{GB}(\tau_{ED}(x_{n_j})),x_{n_i})
\end{align}
where $x_{n_j}$ and $x_{n_i}$ refer to the normal images, 
$\tau_{ED}$ denotes elastic deformation,
$\tau_{GB}$ denotes Gaussian blur,
and $\tau_{LT}$ denotes linear transformation.

\clearpage
\subsection{Hyperparameters}
In shape simulation,
we generate one tumor for each simulated brain data, 
as there is usually only one tumor in the brain MRI. 
But we generate $K$ tumors for each simulated liver data, 
where $K \in \{1,2,...,15\}$,
since there is often more than one tumor for liver in the real situation.
In texture simulation,
the normalise ratio in linear transformation $r$ is in range $(1.0,3.0)$ for brain tumor and $(\frac{1}{8},\frac{1}{2})$ for liver tumor.

\section{Experiment Details}
\subsection{Baseline Methods}
We compare against the state-of-the-art self-supervised learning approaches with official codes available.
In general, these approaches can be cast into two categories,
namely, single-stage ({\em{i.e.}~unsupervised anomaly detection}~\cite{marimont2021implicit, Dey2021ASCNetA}) 
and two-stage ({\em{i.e.}~pre-training and fine-tuning}~\cite{Zongwei2019, NEURIPS2020_d2dc6368}) methods. 
In essence, all self-supervised learning approaches differ only in the definition of the proxy tasks,
as described in the following: \\[-10pt]

\begin{itemize}[leftmargin=0.15in]
\setlength\itemsep{4pt}
\item {\bf 3D IF}~\cite{marimont2021implicit}: an auto-decoder is used to learn the distribution of healthy images, 
where implicit representation is used to map continuous spatial coordinates and latent vectors to the voxels intensities. 
The differences between the output restored image and input are treated as anomalies.
\item {\bf ASC-Net}~\cite{Dey2021ASCNetA}: an adversarial-based selective cutting neural network aims to decompose an image into two selective cuts based on a reference distribution deﬁned by a set of normal images.
\item {\bf 3D RPL}~\cite{NEURIPS2020_d2dc6368}: 
a proxy task is defined to locate the query patch relative to the central patch.
\item {\bf 3D Jigsaw}~\cite{NEURIPS2020_d2dc6368}: 
a proxy task is defined to solve jigsaw puzzles on the shuffled input volumes.
\item {\bf 3D Rotation}~\cite{NEURIPS2020_d2dc6368}: 
a proxy task is defined to predict the rotation degree applied on input scans.
\item {\bf Model Genesis}~\cite{Zongwei2019}: 
a proxy task is defined to recover the original data volume, 
from one that undergoes various artificial transformations, such as non-linear, local-shuffling, outer-cutout and inner-cutout.
\end{itemize}

\subsection{Network Architecture}
Our proposed self-supervised learning method is flexible for any 3D encoder-decoder network.
For simplicity, we adopt the commonly used 3D UNet~\cite{iek20163DUL} as the network backbone,
which is the same as in previous work~\cite{Zongwei2019}.
It comprises an encoder and a decoder with skip connections,
the encoder consists of 4 blocks,
each block has two $3\times 3\times 3$ convolution layers followed by a batch normalization layer, 
a ReLU layer and a $2\times 2\times 2$ max-pooling layer. 
Symmetrically, 
each decoder block contains a $2\times 2\times 2$ deconvolution, 
followed by two $3\times 3\times 3$ convolutions with batch normalisation and ReLU.
In the last layer, 
a $1 \times 1 \times 1$ convolution is used to reduce the number of output channels to 3, 
corresponding to the restored normal organ image, restored tumor and predicted mask. \\[-8pt]

\subsection{Training Details}
For all the two-stage methods, we run their methods with the same pretraining datasets as ours. 
At the {\em fine-tuning} stage, 
the pre-trained encoder is used as an initialization for the network,
and a randomly initialized decoder with skip connections for supervised fine-tuning. 
Data augmentation during fine-tuning involves rotations, scaling, Gaussian noise, Gaussian blur, brightness, contrast, gamma and mirroring, as proposed in the nnU-Net~\cite{Isensee2021}.

\subsection{Evaluation Metrics}
\label{sec4.3}
We quantitatively evaluate the segmentation accuracy using Dice overlap coefficient, sensitivity, specificity, and Hausdorff distance $(95\%)$, specifically: 
\begin{align}
    &\mathrm{Dice} = \frac{2TP}{FP+2TP+FN} \\[3pt]
    &\mathrm{Sensitivity} = \frac{TP}{TP+FN}  \\[3pt]
    &\mathrm{Specificity} = \frac{TN}{TN+FP} 
\end{align}
where TP, FP, TN, FN denote true positives, false positives, 
true negatives and false negatives,
and Hausdorff distance is defined as:
\begin{align}
& d_{\mathrm{H}}(X, Y)=\max \left\{\sup _{x \in X} d(x, Y), \sup _{y \in Y} d(X, y)\right\} \label{eq14} \\
& d(x, Y) = \inf _{y \in Y} d(x, y) \label{eq15}
\end{align}
with $X$, $Y$ referring to predicted mask and ground truth respectively, 
and $x$, $y$ are points on $X$ and $Y$. 

\section{Broader Social Impact}
According to the World Health Organization\cite{WHO2020},
cancer is a leading cause of death worldwide,
accounting for nearly 10 million deaths in 2020. 
However, many cancers can be cured by surgery or systemic therapies when detected before they have metastasized. 
Radiography images are commonly used non-invasive methods for visualizing the inner body structures for disease diagnosis, 
which have received increasing attention from both deep learning and medical communities.
An automatic technique for tumor segmentation can support radiologists to deliver key information about the volume, localization, and shape of tumors to aid early detection and treatment.
Since acquiring the expert pixel-wise annotations for tumor segmentation can often be a prohibitively expensive process, 
we target self-supervised representation learning for zero-shot tumor segmentation, 
{\em i.e.} without any requirement of manual annotation.
Our work aims to provide the necessary tools for general 3D tumor segmentation, 
thus assisting physicians and radiologists in their cancer diagnostic processes,
and the experimental results demonstrate that our approach can help reduce the annotation effort.
By performing routine tumor segmentation tasks, our work can help free up radiologists to perform more value-added tasks, playing a vital role in integrated clinical teams to improve patient care.

\section{Visualisation Results}
\begin{figure}[tb]
\hspace{-2pt}
\centering{\includegraphics[width=0.95\columnwidth]{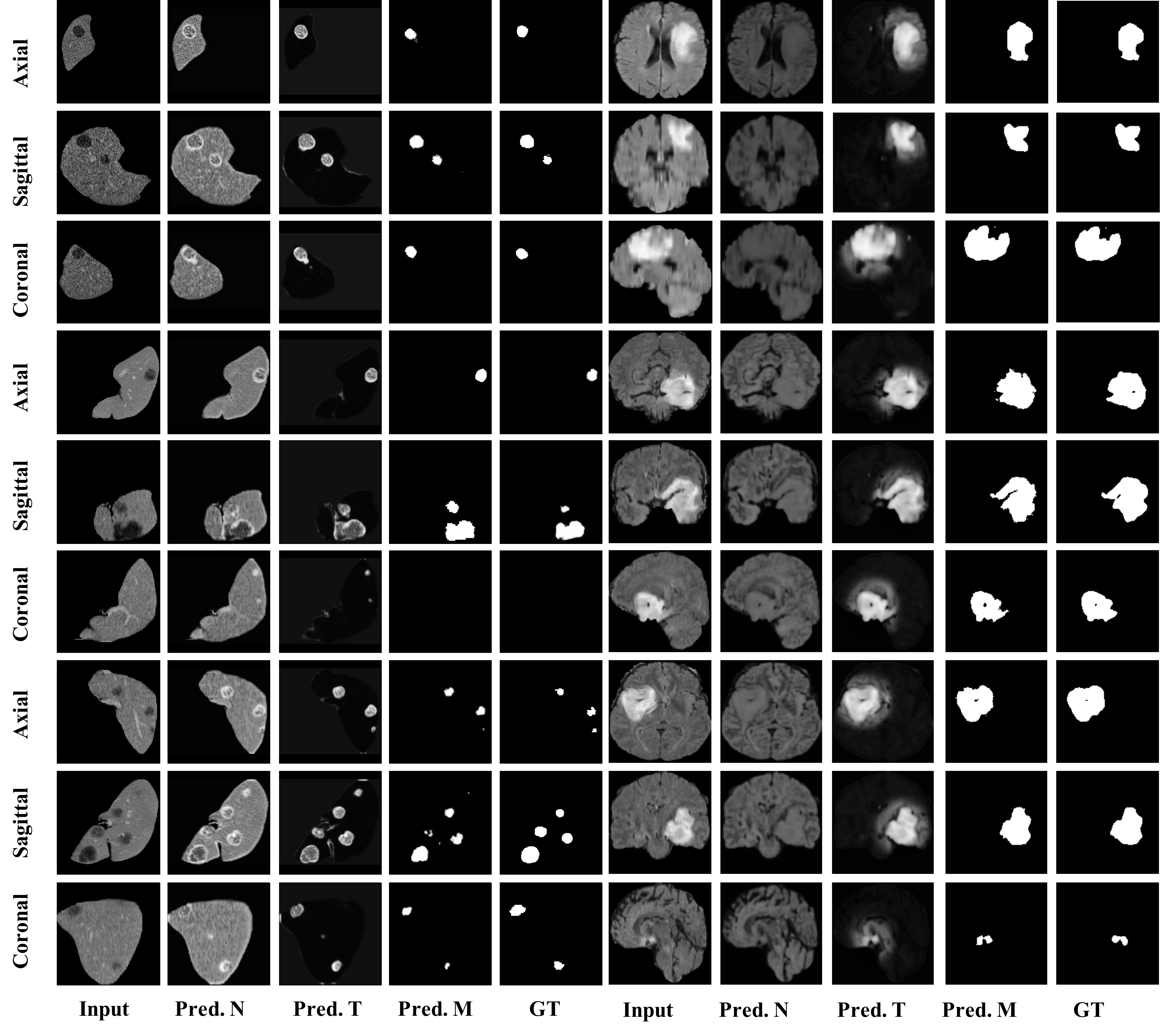}}
\caption{
\textbf{Zero-shot tumor segmentation from our self-supervised model on real tumor data.}
The rows from top to bottom present the slices viewed from axial, coronal, and sagittal views. 
For liver tumor (left) and brain tumor (right), we show: 
Input, Predicted normal image (Pred.N), Predicted tumor image (Pred.T), Predicted mask (Pred.M), Ground Truth (GT).
Note that we do not favor high-quality reconstructions,
what we really care about is the tumor segmentation mask.
}
\label{supple_fig2}
\end{figure}

We show more qualitative results in Figure~\ref{supple_fig2}.

\end{document}